\documentclass[letterpaper, 10 pt, conference]{ieeeconf}  

\IEEEoverridecommandlockouts                              

\usepackage{graphicx} 
\usepackage{amssymb}  
\usepackage{amsmath}
\usepackage[dvipsnames]{xcolor}
\usepackage{verbatim}
\usepackage{xcolor}
\usepackage{booktabs}
\newcommand{\secref}[1]{Sec.~\ref{#1}}
\newcommand{\figref}[1]{Fig.~\ref{#1}}

\definecolor{egolane}{RGB}{77, 80, 206}
\definecolor{otherlane}{RGB}{201, 67, 71}
\definecolor{map}{RGB}{2, 245, 2}

\usepackage{array,multirow}

\title{\LARGE \bf
Robust Monocular Localization in Sparse HD Maps Leveraging Multi-Task Uncertainty Estimation}

\author{K\"ursat Petek$^{*}$, Kshitij Sirohi$^{*}$, Daniel B\"uscher and Wolfram Burgard
\thanks{$^{*}$These authors contributed equally.
All authors are with the Department of Computer Science, University of Freiburg, Germany.
This work was partly financed by the Baden-Württemberg Stiftung gGmbH (perception module) and by the Bundesministerium für Bildung und Forschung (localization method).
}%
}

\begin{document}

\maketitle
\thispagestyle{empty}
\pagestyle{empty}

\begin{abstract}
Robust localization in dense urban scenarios using a low-cost sensor setup and sparse HD maps is highly relevant for the current advances in autonomous driving, but remains a challenging topic in research. We present a novel monocular localization approach based on a sliding-window pose graph that leverages predicted uncertainties for increased precision and robustness against challenging scenarios and per-frame failures. To this end, we propose an efficient multi-task uncertainty-aware perception module, which covers semantic segmentation, as well as bounding box detection, to enable the localization of vehicles in sparse maps, containing only lane borders and traffic lights. Further, we design differentiable cost maps that are directly generated from the estimated uncertainties. This opens up the possibility to minimize the reprojection loss of amorphous map elements in an association-free and uncertainty-aware manner. Extensive evaluation on the Lyft 5 dataset shows that, despite the sparsity of the map, our approach enables robust and accurate 6D localization in challenging urban scenarios using only monocular camera images and vehicle odometry.
\end{abstract}


\section{Introduction}
Despite recent developments in the field of autonomous driving, HD maps remain an indispensable component in modern systems as they provide detailed information on the road infrastructure enabling various applications such as motion planning and enhanced perception. However, to utilize HD maps, it is necessary to accurately determine the vehicle pose within the map.
To solve this localization task, high precision systems generally employ RTK-GNSS systems with integrated IMUs or vehicle odometry in a probabilistic fusion scheme~\cite{jo2011interacting,schreiber2016vehicle}. However, the high cost of equipment and limitations in dense urban environments render these methods limited to research and data generation. Therefore, most deployed autonomous vehicles utilize an HD map that stores information about the environmental elements like lane topology, traffic signs, and traffic lights to localize. 
With the availability of this information, landmark-based localization has gained interest. Such methods consist of a perception module to extract the necessary features from the sensor readings and continuously match them to the available map elements to constrain the vehicle pose~\cite{caselitz2016monocular,cattaneo2019cmrnet}.

Supported by the extensive research and advances in deep learning, the perception modules in modern systems typically consist of a convolutional neural network (CNN) to extract the necessary features from the environment~\cite{poggenhans2018precise},~\cite{pauls2020monocular}. 
Although state-of-the-art CNN architectures can provide a holistic understanding of the environment utilizing sensors such as cameras~\cite{porzi2019seamless,mohan2021efficientps} and LiDARs~\cite{sirohi2021efficientlps}, most CNNs cannot give a reliable confidence estimate or are overconfident about their predictions. This can compromise a localization algorithm's robustness and accuracy.

The task of uncertainty estimation with deep learning extends the standard neural network-based methods to additionally predict the associated uncertainty or confidence in the prediction. Popular uncertainty estimation methods primarily utilize the sampling-based methods~\cite{gal2016dropout}, which are computation and time-intensive. While the research for sampling-free methods is gaining interest, current approaches focus on predicting the uncertainties for a single task like classification~\cite{sensoy2018evidential} or regression~\cite{amini2019deep}. In contrast, an overall perception system of autonomous vehicles consists of various tasks, like segmentation and detection.

In this paper, we aim to solve the localization problem in challenging urban scenarios with a low-cost sensor setup and extremely sparse HD maps containing only lane borders and traffic lights. We present a novel monocular camera-based localization system that leverages the uncertainty estimations of our proposed multi-task perception module. The perception module simultaneously predicts the uncertainties associated with semantics of the lane and with bounding box parameters of the traffic lights in a single pass. 

The main contributions of this paper are: 1) a novel pose graph localization system robust in challenging scenarios by exploiting predicted uncertainties, 2) a multi-task uncertainty-aware perception module capable of simultaneously predicting semantic and regression uncertainties in a sampling free fashion, 3) a novel association-free and differentiable cost map generation module guided by prediction uncertainties. 

We demonstrate the performance gain by incorporating uncertainties in our localization method by evaluating on the challenging Lyft 5 dataset~\cite{kesten2019lyft}.
\begin{figure*}
\centering
\includegraphics[width=1.0\textwidth]{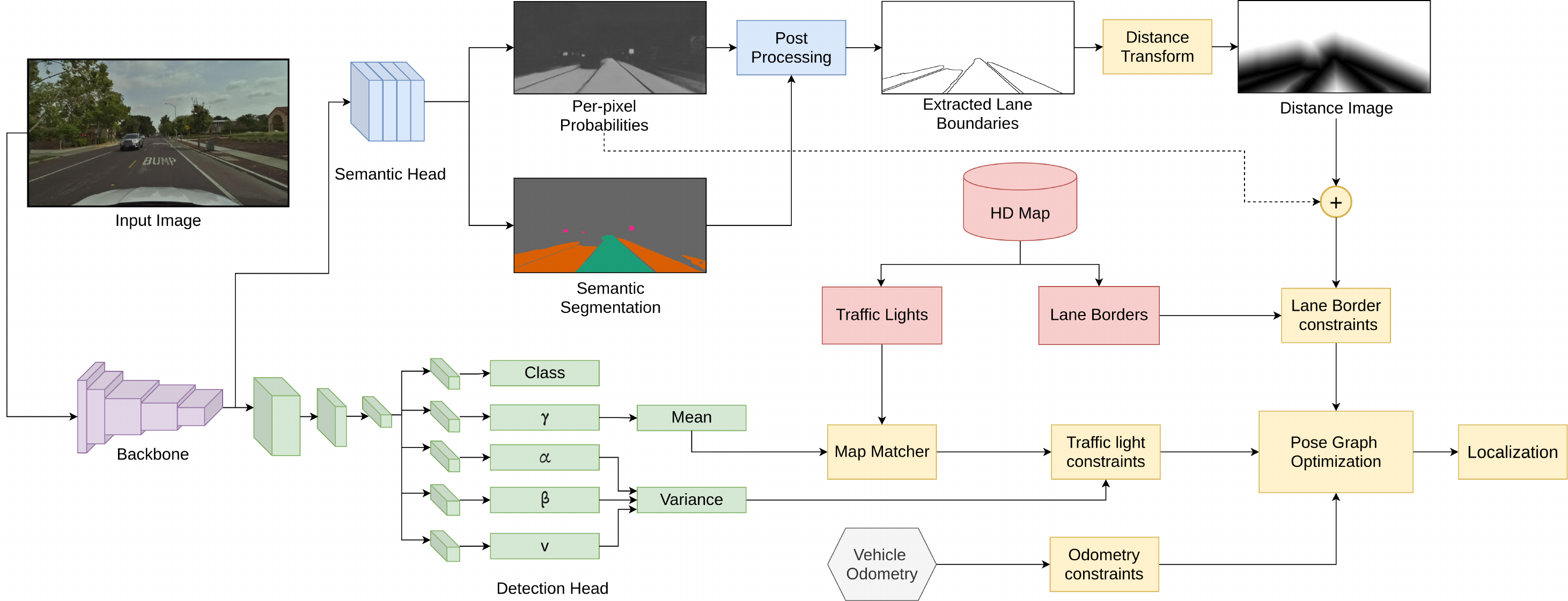}
\caption{Overview of our method. The input image is fed into a multi-head uncertainty-aware network with separate semantic (blue) and detector heads (green) to predict semantic and bounding box uncertainties together with their respective tasks. The predicted semantic probabilities, derived from uncertainties, are used to extract lane borders in the post-processing step. A distance transform is applied to these boundaries, followed by weighting with semantic probabilities to create a cost map. Traffic lights are matched to their corresponding detections by the map matcher. Finally, both perception constraints are set up and fed into the pose graph optimization along with odometry constraints for robust localization.}
\label{fig:network}
\vspace{-0.5cm}
\end{figure*}

\section{Related Work}
\subsection{Localization}

Recent works on localization mainly differ between the utilized map elements, the employed sensor setup, and the perception module. GNSS-based approaches fuse short-term accurate proprioceptive sensor information, e.g., vehicle odometry or IMU, with long-term accurate GNSS information in a tightly coupled manner~\cite{jo2011interacting,schreiber2016vehicle}. However,  the limitations of GNSS systems in dense urban scenarios affect the reliability of these methods. Other methods employ LiDAR maps for accurate localization~\cite{caselitz2016monocular,cattaneo2019cmrnet}, but the reliance on the costly sensor or memory-intensive dense LiDAR maps impacts the scalability of such methods. Other approaches include specialized methods for extracting  landmarks such as poles~\cite{schaefer2019long,kummerle2019accurate}, lane markings~\cite{poggenhans2018precise,schreiber2016vehicle}, and facades~\cite{kummerle2019accurate}. Although being accurate, these methods require specialized detectors and mapping procedures for reliable localization. 

Hence, deep learning-based methods have gained importance to render the localization flexible and scalable with the availability of additional data. Radwan et al.~\cite{radwan2018vlocnet++} propose to use a fully learning-based visual localization method that predicts the pose difference between consecutive images and the global pose of each frame. Pauls et al.~\cite{pauls2020monocular} introduces a hybrid monocular localization method combining the advantages of deep learning and classical approaches. However, they use a pre-implemented network without adaptations according to the needs of localization tasks and the method is unaware of the inherent network uncertainties, rendering it unreliable in challenging environments.

\subsection{Uncertainty Estimation}
Uncertainties are classified into aleatoric (data) uncertainty to quantify the noise in data and epistemic (model) uncertainty to quantify uncertainty in model prediction due to lack of training data or insufficient knowledge of the model~\cite{kendall2017uncertainties}. While it is possible to derive data uncertainty from data statistics or learn it with a network, it is harder to predict epistemic uncertainty due to the intractability of exact Bayesian inference for neural networks. To this end, most methods employ the popular sampling-based Monte Carlo (MC) dropout technique~\cite{gal2016dropout} and Bayesian neural networks (BNNs). For example, methods such as~\cite{mukhoti2018evaluating,huang2018efficient}, and~\cite{harakeh2020bayesod,kraus2019uncertainty} employ modified versions of MC dropout to predict per pixel semantic and bounding box regression uncertainties, respectively.


In bounding box uncertainty estimation, approaches such as~\cite{he2019bounding,le2018uncertainty} generally predict the aleatoric uncertainty for each of the bounding box parameters using a modified loss function by taking an extra variance term into account.


Sensoy et al.~\cite{sensoy2018evidential} introduce a sampling-free method called deep evidential learning to quantify the classification uncertainty by making the network collect evidence to predict higher-order prior distribution parameters. In this context, evidence denotes the magnitude of support the network predicts in favor of classifying a sample to a particular class. The approach in~\cite{amini2019deep} proposes evidential deep learning for the task of monocular depth estimation, which is a regression task. Both approaches show the strength of evidential deep learning by providing comparable or superior results to most sampling-based methods. Hence we utilize evidential deep learning to simultaneously predict semantic segmentation and bounding box detection uncertainties in a single pass.

\section{Technical Approach}
Our localization method consists of an uncertainty-aware perception module,
a differentiable cost-map generator, a map matcher and a pose graph optimization module (see~\figref{fig:network}).
The perception module incorporates a semantic head for spatially unconstrained map elements, i.e. driveable areas,
and a bounding box detection head for map elements with a finite extent like traffic lights.

The segmentation outputs are processed along with the estimated uncertainties to create a differentiable cost-map in the image plane which,
in combination with the corresponding map elements,
provides the lateral constraints for the camera pose from lane borders.
The detection head detects traffic lights represented as bounding boxes together with the uncertainties associated with each parameter. 

In the next step, the map matcher associates each potentially visible traffic light from the map with its counterpart from the detection module.
The bounding boxes and the reprojections of potentially visible traffic lights serve as the inputs to compute the cost term.
Traffic light constraints are set up based on these associations to penalize the point-to-point pixel distance between the instances and the reprojected traffic lights.

In the final step, the sliding-window pose graph optimization problem combines the constraints from the traffic lights and lane borders with odometry constraints to robustify the method and overcome per-frame failures in the perception module.
In the end, optimization provides the most recent pose $\boldsymbol{p}^*$ as the localization result.

\subsection{Perception Module}

We use a convolutional neural network for the perception module.
The architecture consists of a shared EfficientNet-B3~\cite{tan2019efficientnet} backbone with a feature pyramid network on top~\cite{lin2017feature}.
The backbone learns features at multiple scales, utilized by separate semantic and detection heads.

\subsubsection{Uncertainty-Aware Semantic Segmentation Head}
Our semantic segmentation head is a modified version of the semantic head proposed in~\cite{porzi2019seamless}, which takes features at multiple scales and upscales them to a common scale followed by concatenation. We modify the semantic head by replacing the softmax at the end of the network with ReLU, which serves as an evidence signal of the model.

The evidential deep learning method proposes to estimate high order conjugate priors over the network output distribution to estimate the classification uncertainties~\cite{sensoy2018evidential}. We use the Dirichlet distribution as the prior for multinomial classification prediction per pixel, which is parameterized by $N$ parameters $\alpha$ = $[\alpha_{1},..,\alpha_{N}]$ and the network is trained to predict $\alpha_{i}$ for each class $i$ of total $N$ classes. For semantic segmentation, the network predicts $\alpha$ for every pixel of the image. 

We utilize the sum of squares form of the loss $L(\zeta)_i$ to penalize the misclassified pixels and the Kullback-Leibler (KL) divergence loss $\mathcal{L}^\text{KL}_{i}$ to predict high uncertainties for low evidence predictions for pixel $i$, as described by~\cite{sensoy2018evidential}. Moreover, as our application is semantic segmentation, we formulate the overall semantic loss as
 \begin{equation}
\mathcal{L}_{sem} = \sum_{w=1}^W \sum_{h=1}^H {L(\zeta)}_{{w,h}} + \lambda_{s}  \sum_{w=1}^W \sum_{h=1}^H {L}^{KL}_{w,h},
\end{equation}
where $W$ and $H$ are the width and height of the image, respectively, and $\lambda_{s}$ is the annealing coefficient. We use $\lambda_{s} = \text{min}(1.0, t/4)$, where $t$ is the ratio of the current iteration number and the total iterations per epoch.

\subsubsection{Uncertainty-Aware Object Detection Head}
For the detection head, we use a modified Faster-RCNN network to predict the class, bounding box parameters, and additional three parameters required for uncertainty estimation. We define the bounding box by the parameters ($x_\text{min}$, $y_\text{min}$, $x_\text{max}$ and $y_\text{max}$). Hence, the estimation of the bounding boxes is formulated as a regression problem. The aim is to estimate the mean $\mu$ and the associated variance $\sigma^2$ for each of the four bounding box parameters.

As proposed by~\cite{amini2019deep}, we utilize Normal-Inverse-Gamma (NIG) distribution as conjugate prior for evidential regression learning. NIG is defined by four parameters, $\gamma,\alpha,\beta$ and $\upsilon$. To this end, we extend the Faster-RCNN~\cite{ren2015faster} network to have four separate branches to predict these parameters, as depicted in~\figref{fig:network}. In this formulation the mean value $\mu$ is given by $\gamma$. The associated aleatoric uncertainty is calculated as $U_\text{a} =\beta/(\alpha -1)$  and the epistemic uncertainty as $U_\text{e} = U_\text{a}/ \upsilon$.


We train our object detection head with the negative log-likelihood loss  $\mathcal{L}_\text{NLL}$ for maximizing the model evidence or correct predictions and a regularizer term for penalizing errors scaled by the evidence $\mathcal{L}_\text{R}$~\cite{amini2019deep}. With the addition of the common objectness score loss $\mathcal{L}_\text{os}$ and the object proposal loss $\mathcal{L}_\text{op}$~\cite{ren2015faster}, the total detection loss is defined as:
\begin{equation}
\mathcal{L}_\text{det} = \mathcal{L}_\text{NLL} + \lambda_\text{det} \mathcal{L}_\text{R} + \mathcal{L}_\text{os} + \mathcal{L}_\text{op}
\end{equation}
The authors of~\cite{amini2019deep} suggest using the scaling factor $\lambda_\text{det}=0.01$ for the task of depth regression. However, the network tends to predict overconfident estimations with this value for the task of bounding box regression. Thus we have used $\lambda_\text{det}=0.04$.

Finally, the overall loss is defined as
$\mathcal{L} = \mathcal{L}_{sem} + \lambda \mathcal{L}_\text{det}$,
where we use a scaling factor of $\lambda=15$, since $\mathcal{L}_{sem}$ and $\mathcal{L}_\text{det}$ have different scales.

\subsection{Differentiable Cost Map Generation}
The availability of lane topologies for many roads in HD maps makes their usage highly relevant for the task of localization in autonomous driving~\cite{homayounfar2019dagmapper}. This task is composed of detecting the lane borders and matching them to their counterparts in the HD map. As such, we want to detect any consistent longitudinal feature on the road as lane borders, such as lane markings, road boundaries, and parking zones.

\begin{figure}
\setlength\tabcolsep{2pt}
\centering
\begin{tabular}{|c|c|c|}
 \hline
 {}&{Softmax} &
 {Ours} \\\hline
 \rotatebox{90}{Probabilities} &\includegraphics[width=1.4in]{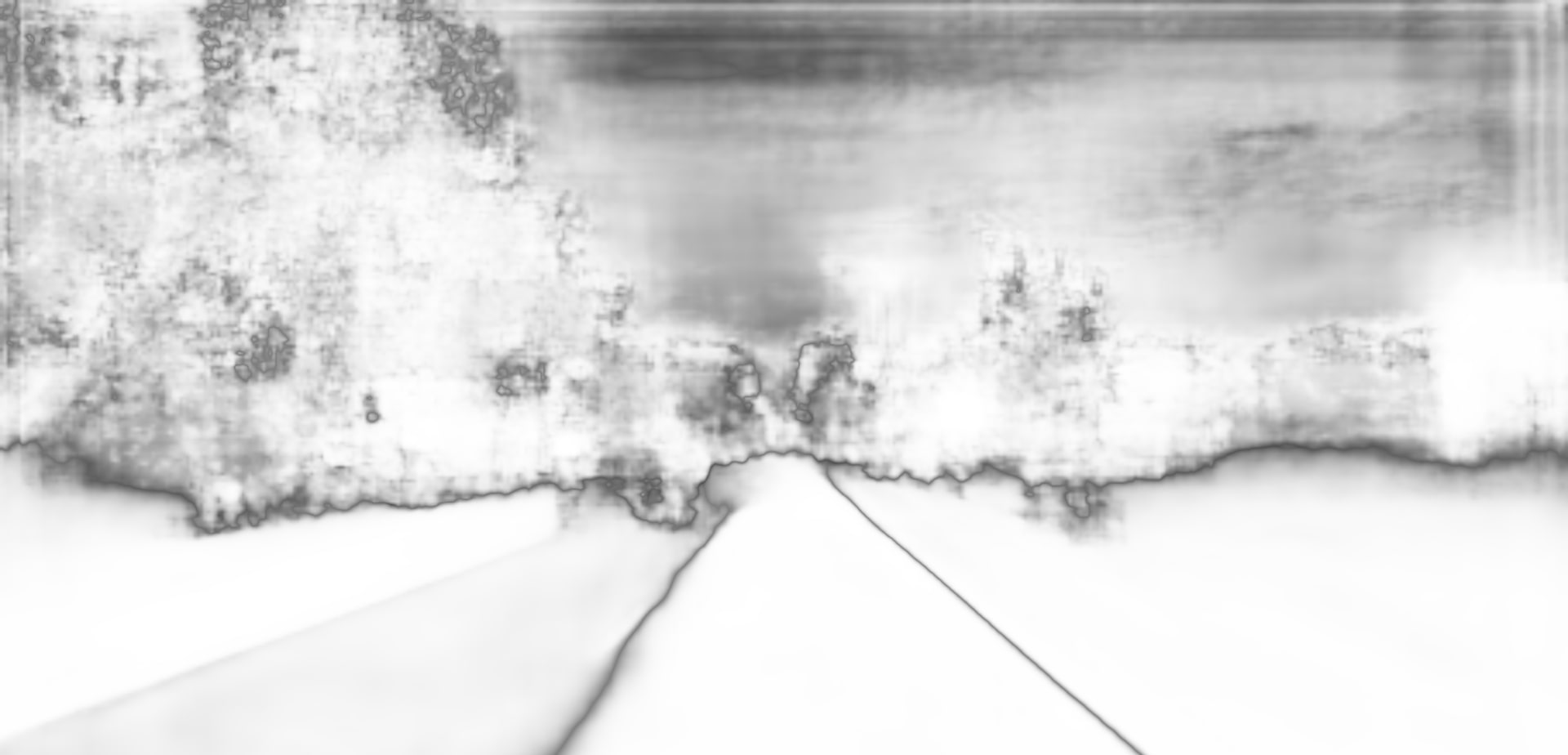} &
 \includegraphics[width=1.4in]{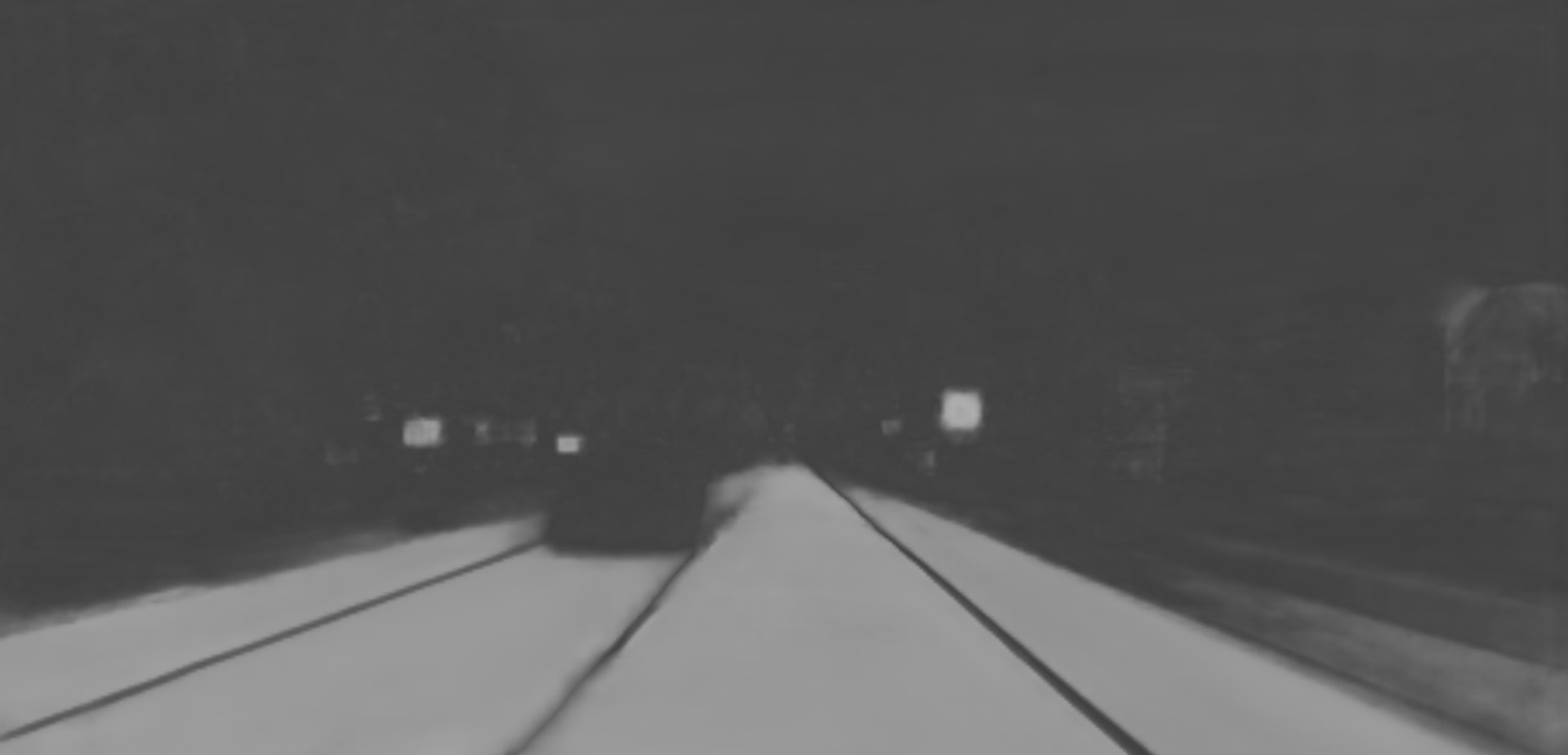} \\\hline
 \rotatebox{90}{Boundaries} & \includegraphics[width=1.4in]{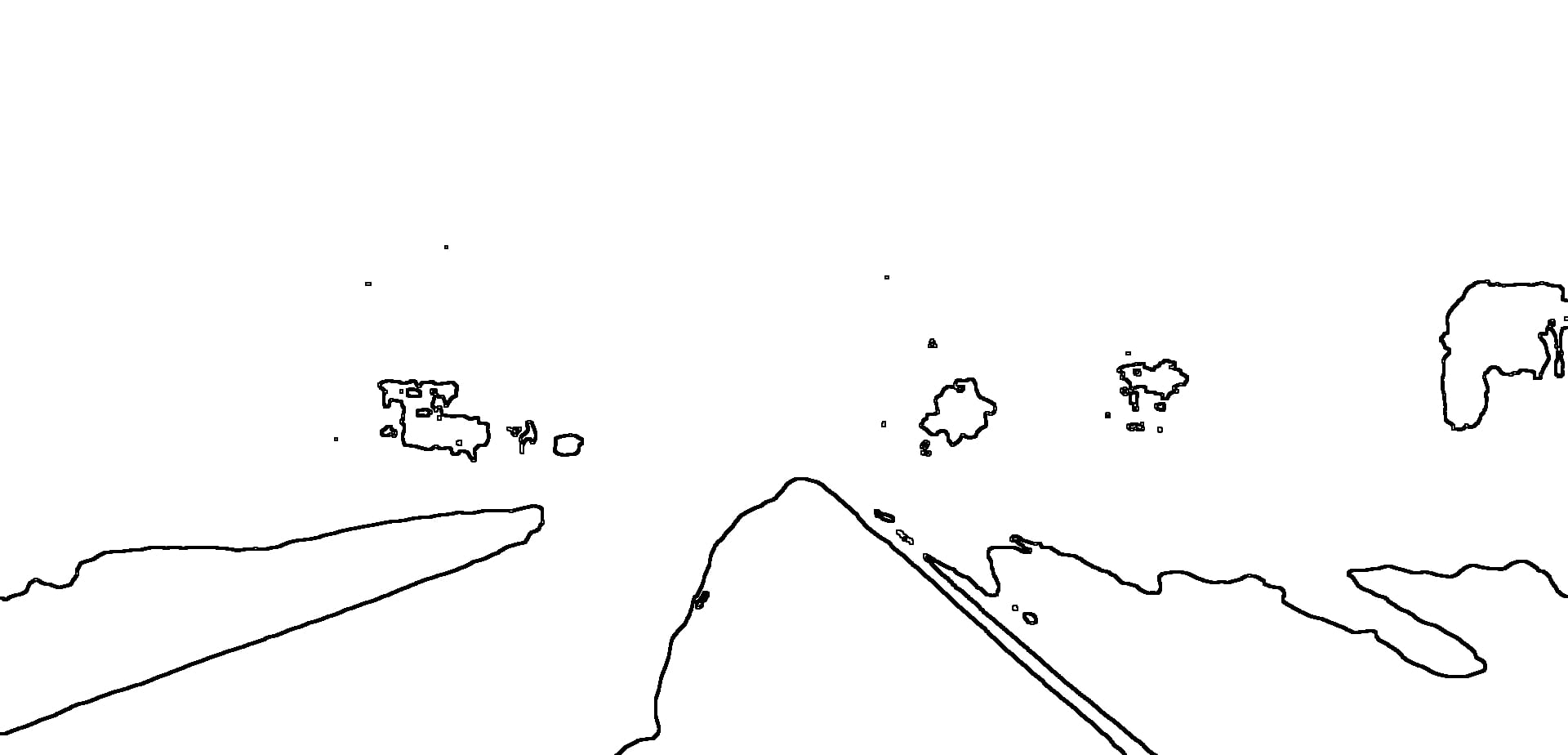} &
 \includegraphics[width=1.4in]{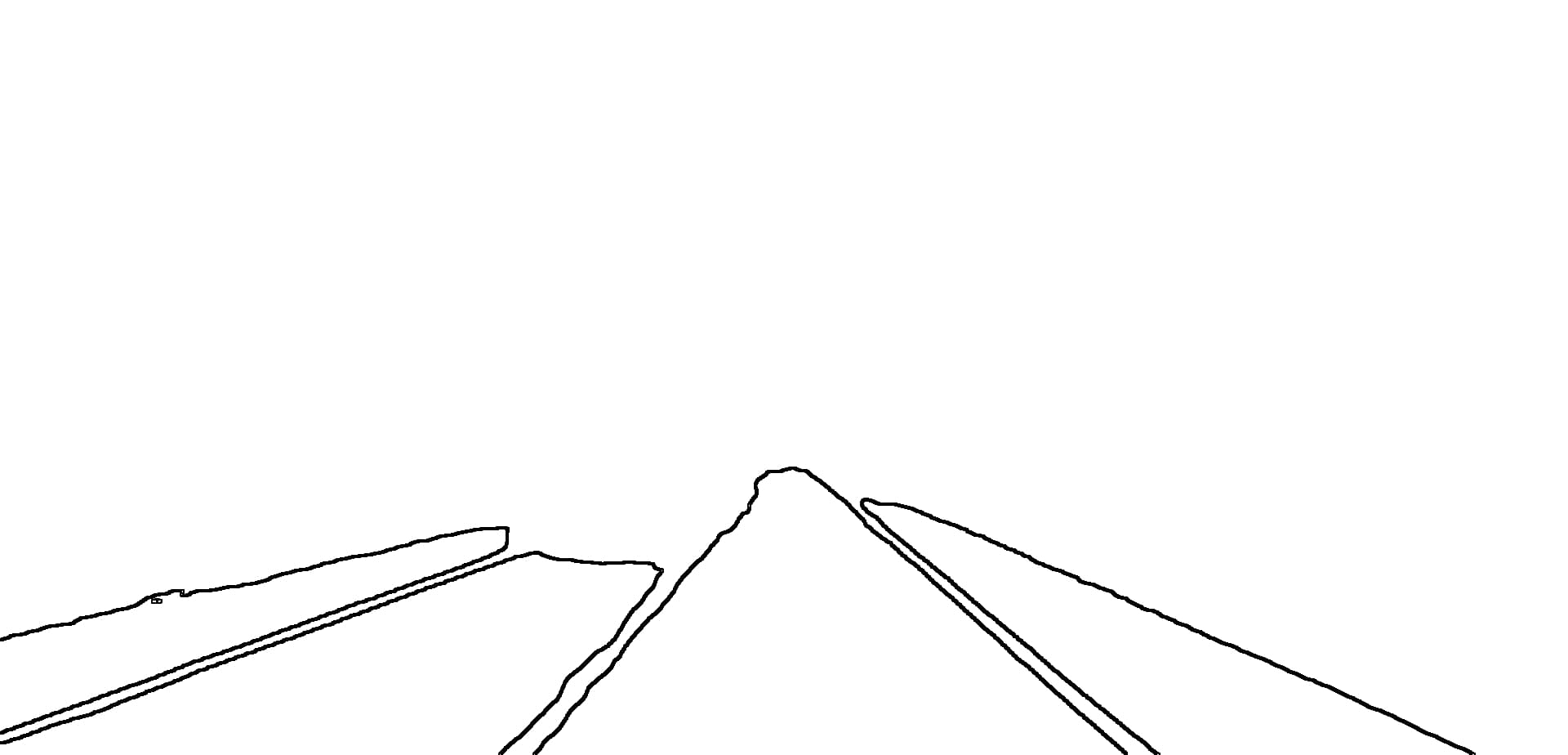} \\\hline
\end{tabular}
\caption{Comparison of the predicted probabilities (upper) and borders extracted using probabilities (lower) between a simple softmax (left) and our uncertainty-aware semantic segmentation (right). Light areas represent high and dark regions represent low probability. The softmax tends to overestimate the probabilities for all regions, whereas our calibrated probability estimation assigns low values to non-lane areas and leads to better border estimations.}
\label{fig:softmax}
\end{figure}

The semantic head of our network provides segmentation for direct drivable and alternative drivable areas on the road with associated uncertainty values. Due to our well-calibrated uncertainty predictions (see Section~\ref{sec:unc}) the probability map $P_{s}$ consistently yields high uncertainty values for continuous longitudinal distortions on and off the road. Thus, for extracting the aforementioned longitudinal features, we simply threshold the uncertainty map $I_{\mathrm{unc}}$ with Otsu's method that maximizes the separability of the uncertainty values~\cite{otsu1979threshold} and mask out all non-lane classes to obtain the fully segmented lane borders. The result is an image $I_\text{lb}$ containing only the lane borders depicted by boundaries in \figref{fig:softmax}.

We optimize an error metric to constrain the vehicle pose to penalize the mismatch between the extracted lane borders and the lane topology map, reprojected onto the image. However, any direct association between the detected lane border pixels and reprojected map elements will be imperfect, since lane borders do not constrain in the longitudinal direction. To overcome this challenge, we apply a distance transform on the detected lane borders as proposed in~\cite{pauls2020monocular}. The distance transform yields a cost-map $\boldsymbol{C}_{s}$ with each pixel containing the euclidean distance to the closest lane border in pixel space.

In the final step, we overlay $\boldsymbol{C}_{s}$ with the weighted probability map and apply a bi-cubic interpolation to obtain the final differentiable cost map $\boldsymbol{C}_{\mathrm{unc}}$. The probabilities show a smooth transition from the lane segments towards the approximate centerline of segmented lane borders yielding additional gradients. This allows for considering the uncertainties of the perception module throughout the whole extent of the lane borders.

\subsection{Map Matching}
The map matching step aims to associate the potentially visible set of reprojected map elements $\boldsymbol{X}^\mathrm{tl}$,
the traffic lights, to the detections provided by the detection head of our perception module. First, we compute the per-pixel distance between each reprojected traffic light center and each detected bounding box center. This distance serves as a quality measure for each potential association. Second, due to the reliability of the map, the reprojected traffic lights are associated with the closest detections. Thus, even an erroneous pose yields correct associations as minor errors have almost no impact on the position of reprojections in the image plane for distant regions. Our perception module is capable of detecting traffic lights from a far distance of approximately 50m. Hence, we apply the map matching as early as possible and keep the correct initial associations until the vehicle moves past the traffic lights under consideration.

\subsection{Sliding-Window Pose Graph Optimization}
Due to the missing redundancy of map elements and the potential per-frame failures in the perception module in highly challenging scenarios, we choose to design a robust sliding-window pose graph optimization method~\cite{grisetti2010tutorial}. This method optimizes $N$ poses simultaneously, constrained by the detected features and the corresponding map elements. In order to obtain the final state vector $\boldsymbol{p}^{*}$, we optimize the cost function $J = J^{\mathrm{o}} + J^{\mathrm{lb}} + J^{\mathrm{tl}}$, accounting for the lane borders ($\mathrm{lb}$), traffic lights ($\mathrm{tl}$) and the odometry ($\mathrm{o}$):

\begin{equation}
    \boldsymbol{p}^{*} = \underset{\boldsymbol{p}}{\arg \min}
    \sum_{i \in \{\mathrm{lb}, \mathrm{tl}, \mathrm{o} \}}
    \sum_{k=1}^{N}
     J^{i}\left(\boldsymbol{p}_{k}, \boldsymbol{z}^{i}_{k}, \boldsymbol{m}\right),
\end{equation}
where $\boldsymbol{z}^{i}_k$ are the detections of the measurement class $i$ for pose $\boldsymbol{p}_k$ and $\boldsymbol{m}$ is the semantic HD map. This cost function can be further split into its error terms $\boldsymbol{e}_k^{i}$ and the corresponding information matrix $\boldsymbol{\Omega}^{i}_k$.

\begin{equation}
\label{eq:pose_graph_all}
\begin{split}
    \boldsymbol{p}^{*} = &\underset{\boldsymbol{p}}{\arg \min}
    \sum_{i \in \{lb, tl \}}
    \sum_{k=1}^{N} \rho
    \left(
        \boldsymbol{e}_{k}^\mathrm{i,T}
        \left(
        \boldsymbol{p}_k
        \right)
        \boldsymbol{\Omega_{k}^{i}}
        \boldsymbol{e}_{k}^{i}
        \left(
        \boldsymbol{p}_k
        \right)
    \right) \\ & +  
    \sum_{k=1}^{N-1} 
    \left(
        \boldsymbol{e}_{k}^{\mathrm{o,T}}
        \left(
        \boldsymbol{p}_k, \boldsymbol{p}_{k+1}
        \right)
        \Omega_{k}^\mathrm{o}\boldsymbol{e}_{k}^{\mathrm{o}}
        \left(
        \boldsymbol{p}_k, \boldsymbol{p}_{k+1}
        \right)
    \right),
\end{split}
\end{equation} where
$\rho(x) = \mathrm{log}(1+x)$ denotes the Cauchy function, which robustifies the method by remapping the loss values via a logarithmic projection and effectively lowers the impact of outliers in the optimization.

The error term related to the lane borders, $\boldsymbol{e}_k^{\mathrm{lb}}$, is obtained by reprojecting the potentially visible map elements directly into the uncertainty cost map $\boldsymbol{C}_{\mathrm{unc}}$, using the forward pinhole camera model $f_{\mathrm{cam}}$ and the pose $\boldsymbol{p}_k$:
\begin{equation}
    \boldsymbol{e}_{k}^{\mathrm{lb}}  = \boldsymbol{C}_{\mathrm{unc}}
    \left(
    f_{\mathrm{cam}}
    \left(
    \boldsymbol{p}_k^{-1} \boldsymbol{X}^\mathrm{lb}
    \right)
    \right),
\end{equation}
where $\boldsymbol{X}^\mathrm{lb}$ denotes the set of all lane border point positions under consideration.

The second error term, $\boldsymbol{e}_k^{\mathrm{tl}}$, directly penalizes the pixel distance between the bounding box and the associated traffic light points:
\begin{equation}
    \boldsymbol{e}_{k}^{\mathrm{tl}}  =
    \boldsymbol{X}_k^{\mathrm{bb}} -
    f_{\mathrm{cam}}
    \left(
    \boldsymbol{p}_k^{-1} \boldsymbol{X}^\mathrm{tl}
    \right),
\end{equation}
where $\boldsymbol{X}^{\mathrm{bb}}$ denotes the pixel positions of the set of traffic lights detected in frame k represented as bounding boxes (bb).
Due to the finite extents of traffic lights, a direct association is possible and no additional cost map has to be generated. This error term is differentiable and can directly be incorporated into the optimization problem.
The probabilities estimated by our uncertainty-aware perception module are considered in the information matrix.

Finally, any displacement on consecutive poses $\boldsymbol{p}_k$ and $\boldsymbol{p}_{k+1}$ imposed by the optimizer that deviates from the odometry  $\boldsymbol{\Delta}_{k\rightarrow k+1}$ is penalized by
\begin{equation}
    \boldsymbol{e}_k^\mathrm{o}\left(
    \boldsymbol{p}_k, \boldsymbol{p}_{k+1}
    \right) = \boldsymbol{p}_k^{-1}\boldsymbol{p}_{k+1} - \boldsymbol{\Delta}_{k\rightarrow k+1}^{\mathrm{meas}}.
\end{equation}

With the final cost function being set up, we can now localize the vehicle within the map by optimizing this overall cost. Alternatively, this method can be executed in a single frame setting by dropping the odometry constraints. This, however, is only done for evaluation purposes (see~\secref{sec:single_image_localization}).

\section{Experimental Evaluation}


\begin{figure*}
\setlength\tabcolsep{2.5pt}
\centering
 \includegraphics[width=2.25in]{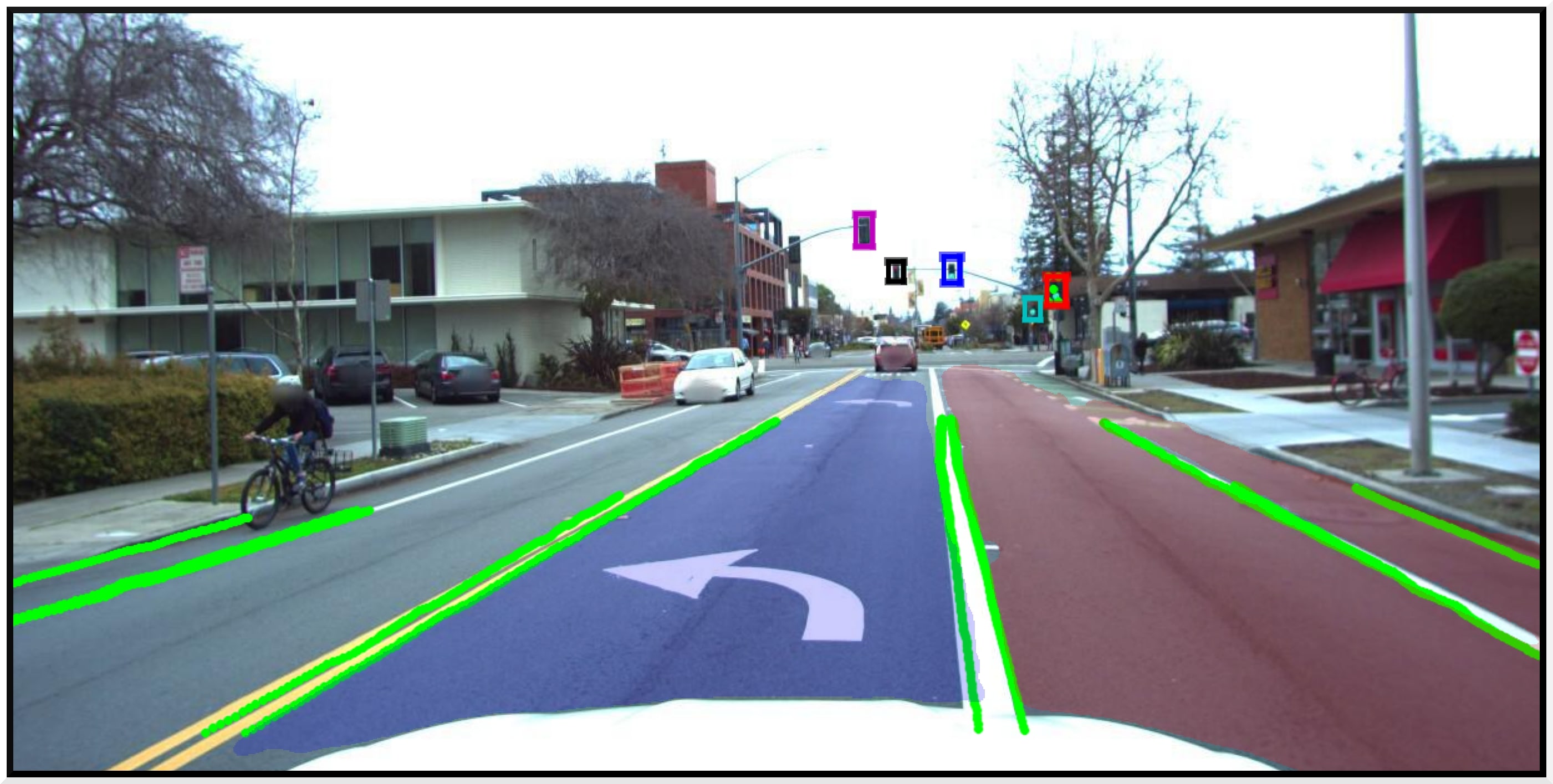}
 \includegraphics[width=2.25in]{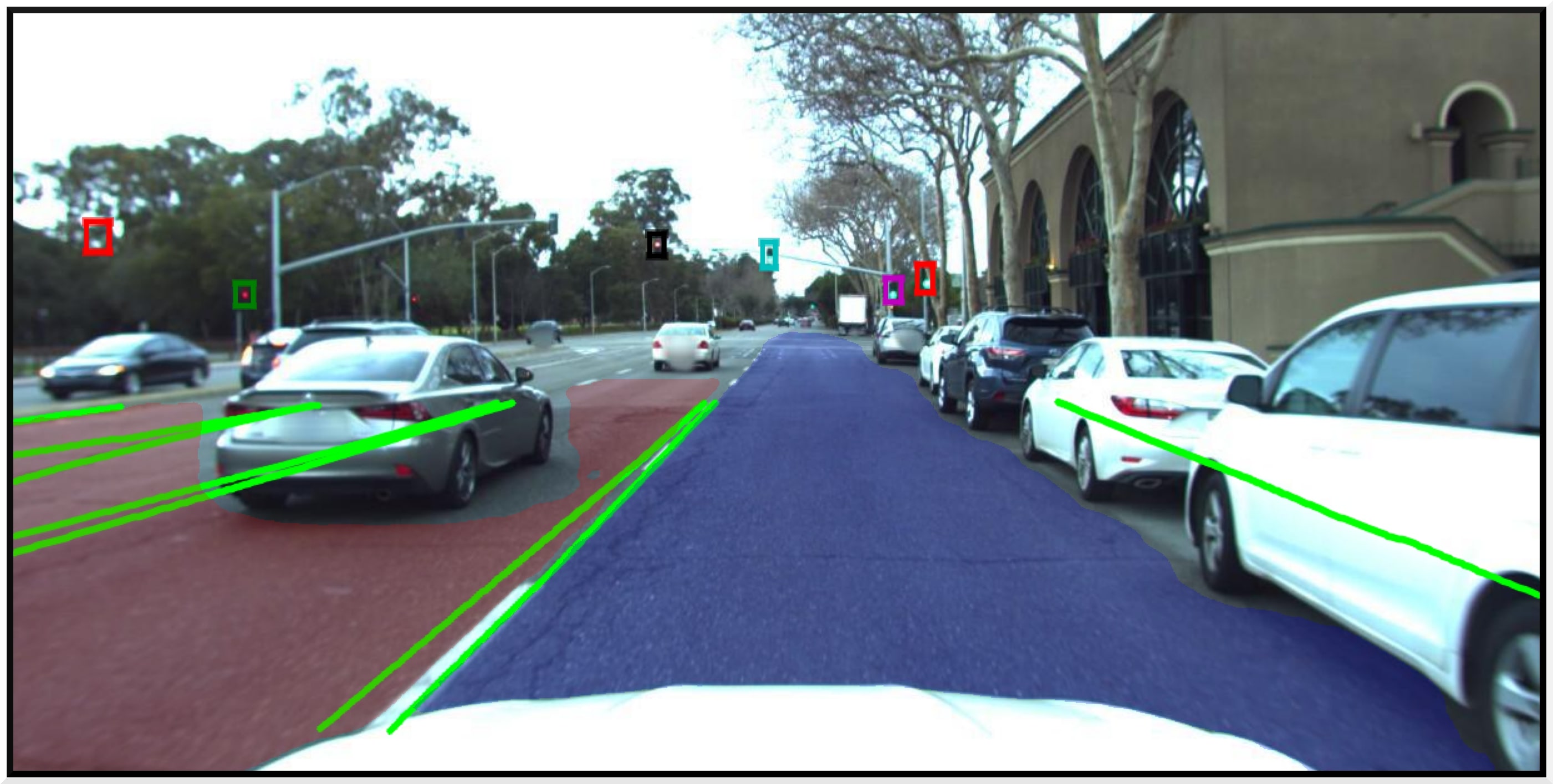}
 \includegraphics[width=2.25in]{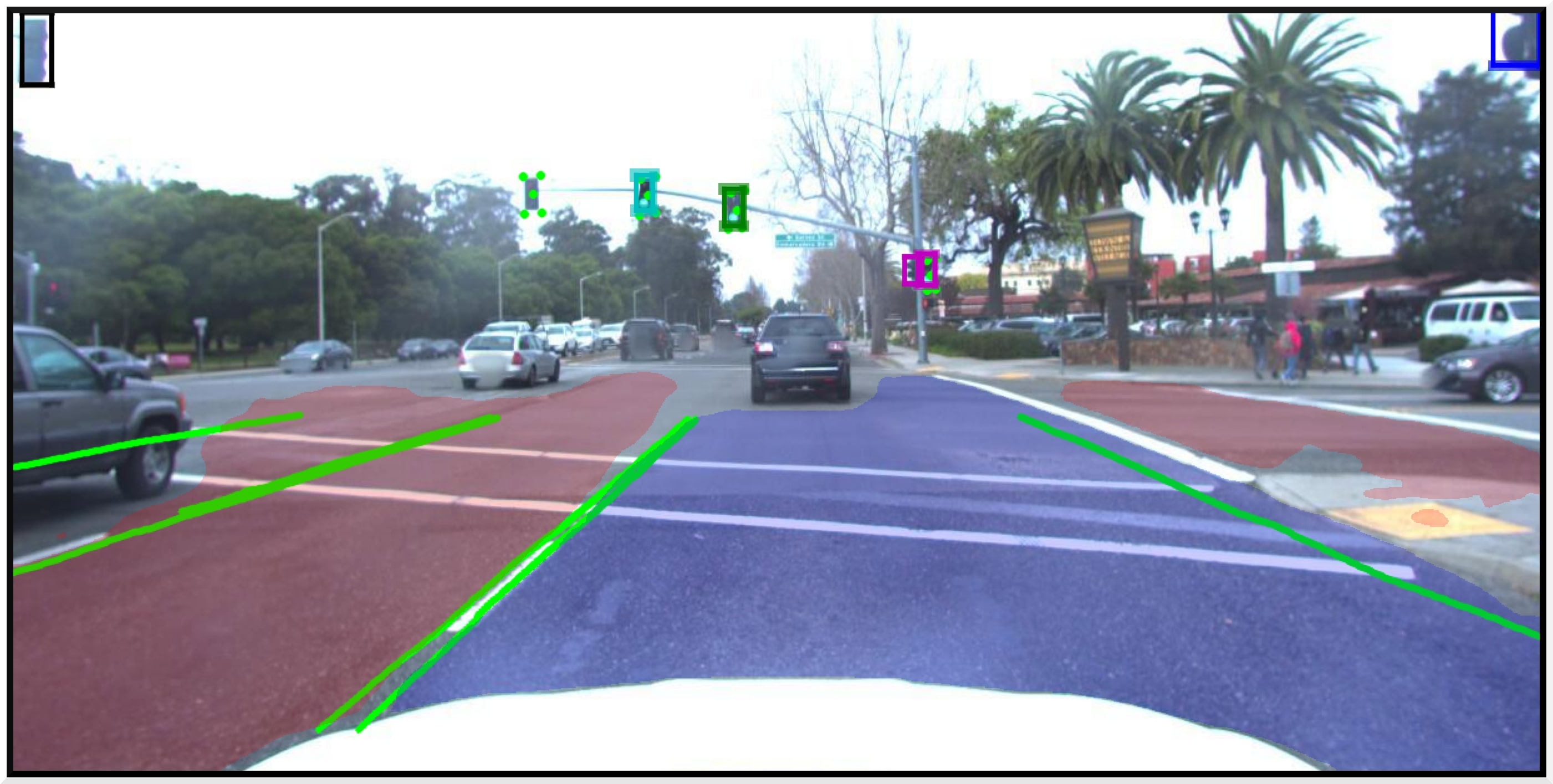}\\
\textcolor{egolane}{$\blacksquare$} direct drivable area\hspace{1em}
\textcolor{otherlane}{$\blacksquare$} alternative drivable area\hspace{1em}
\textcolor{map}{$\blacksquare$} projected map elements
\caption{Reprojected map elements based on our localization results in three different scenarios. Despite occlusions and challenging intersections, our method yields accurate localization results even with a small set of lane borders or traffic lights. The semantics of lanes and bounding boxes are predicted by the perception network. The width of bounding boxes represents the variance predicted for the corresponding edge (best viewed at x4 zoom scale).}
\label{fig:qualit_lyft}
\end{figure*}

\subsection{Dataset}

For evaluating our localization system, we explore the Lyft5~\cite{kesten2019lyft} dataset, which provides 6D localization ground truth and a semantic HD map containing the lane topologies and the sparse instance elements, like traffic lights.
Example images are given in \figref{fig:qualit_lyft}.
However, the scenes are only about 25 seconds long, and the odometry information is also missing. Thus, we order and stitch the scenes together using ground truth poses to create a long continuous sequence with a length of 2.6 km and in a highly populated and challenging urban area. We create the odometry by utilizing the ground truth poses to predict noisy odometry signals according to the velocity-based motion model. The accuracy obtained by the emulated odometry is equally or less precise than modern car odometry systems evident from the longitudinal drift presented in \figref{fig:pgl_quantitative}.

We train our perception network on the bdd100k dataset~\cite{yu2020bdd100k}, containing 70,000 images for training and 10,000 images for validation. We train the semantic head to predict the direct drivable area and alternative driveable area as introduced in~\cite{yu2020bdd100k}. Similarly, the detection head is trained to predict bounding boxes for the traffic light class. To evaluate the performance of our uncertainty estimation, we utilize the validation set, which the network never used during training.



\begin{figure}[]
\centering
\vspace{-0.5cm}
\includegraphics[width=0.45\textwidth]{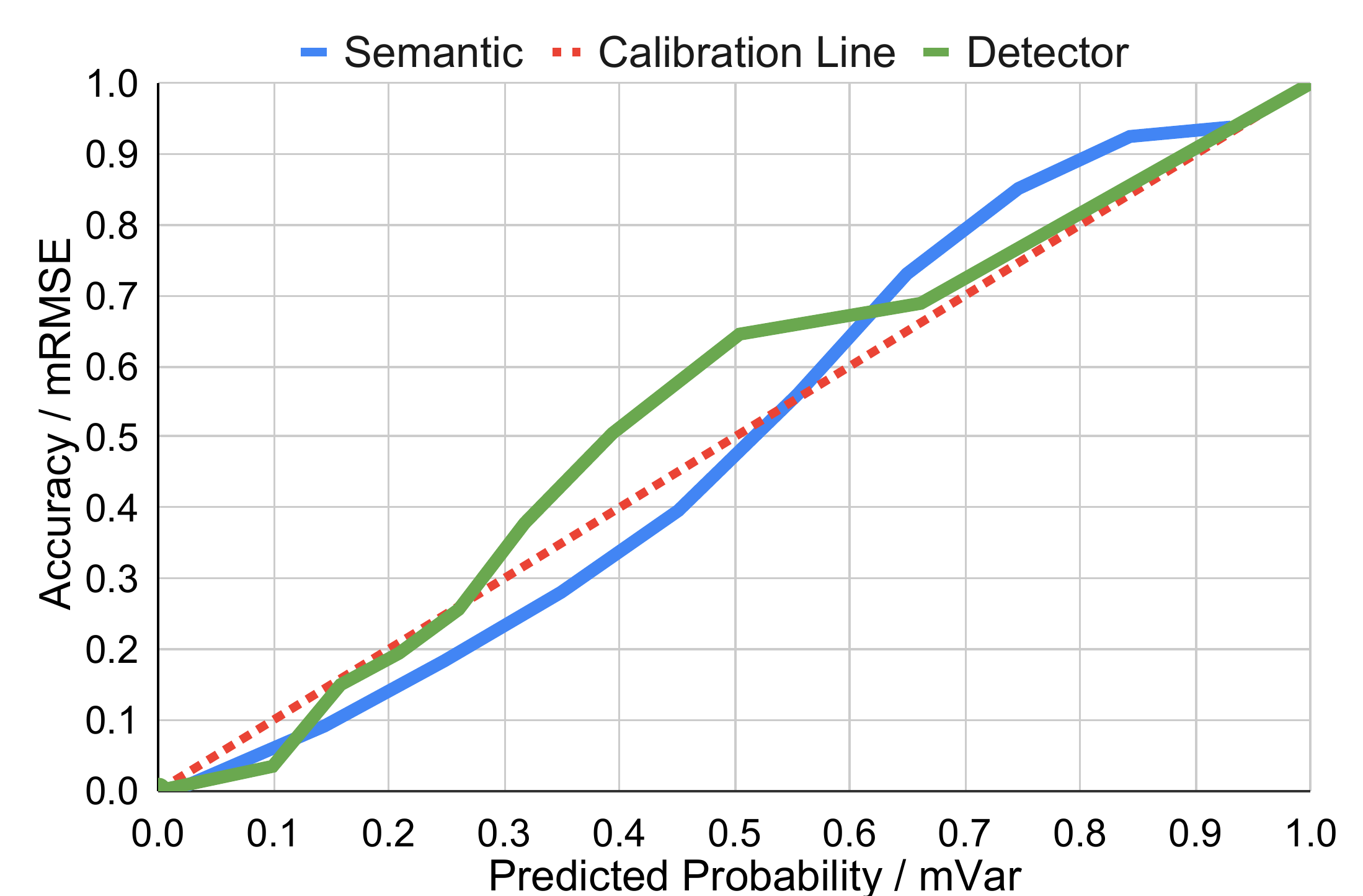}
\caption{Calibration plot for the estimated semantic and detection uncertainties. For semantic segmentation, it shows the accuracy vs. the predicted probability (blue), and for detection, it shows the mRMSE vs. the mVar metric (green). Both uncertainty estimations follow the calibration line (red) in close vicinity, signifying well-calibrated uncertainty estimations.} 
\vspace{-0.5cm}
\label{fig:calibrationcurve}
\end{figure}

\subsection{Uncertainty Estimation}\label{sec:unc}
We evaluate the performance of the semantic segmentation uncertainty estimation to make sure that they are well-calibrated. We report the values of the Expected Calibration Error (ECE). For calculating this metric, the predicted probability axis is divided into $J$ equally spaced bins, and for each bin the average accuracy $\text{acc}(B_{j})$ and average predicted confidence $\text{conf}(B_{j})$ are computed. Then the ECE is given as
\begin{equation}
\text{ECE} = \sum_{j=1}^{J} \frac{n_j}{N} \left| \text{acc}(B_{j}) - \text{conf}(B_{j}) \right|,
\end{equation}
where $n_{j}$ is the number of samples in bin $j$ and $N$ is the total number of samples. The ECE value depicts the deviation from the optimal calibration line. We achieve an ECE of 5.3\%. For bounding box detection, we utilize the Expected Normalized Calibration Error (ENCE)~\cite{levi2019evaluating} which is similar to the ECE. As bounding box detection is a regression task, ENCE reflects the relation between the predicted variance and the Root Mean Square Error (RMSE). The detector achieves an ENCE value of 11.5\% which is reasonable in comparison to the value of 8.5\% for~\cite{kuleshov2018accurate} reported by~\cite{levi2019evaluating} on the KITTI dataset. Note that the lower value signifies better performance in both tasks.

In addition, we show the calibration plot of the accuracy vs. the predicted probability, for semantic segmentation and predicted mean-variance (mVar) per bin vs. (mRMSE) per bin~\cite{levi2019evaluating} for bounding box detection, see~\figref{fig:calibrationcurve}. The desired result is a response that is close to the $y=x$ line. We observe that both tasks follow the calibration line in close vicinity and hence provide meaningful and usable uncertainties. 


\subsection{Single Image Localization}

\label{sec:single_image_localization}
To showcase how uncertainties help improve the accuracy and reliability of the localization method, we evaluate our approach by using a single frame for localization. First, we add translational and a rotational noise, sampled from three different settings of a uniform distribution, onto the ground truth pose to obtain a distorted pose. Second, we initialize the localizer with this distorted pose and relocalize the camera within the lane using the lateral constraints.
Here, we omit the odometry constraints to cancel out their impact on the localization result.

We compare two settings, one with the distance transform directly applied on the predicted lane classes without utilizing the uncertainties (D), and the other using the uncertainties for segmenting all lane/road borders (D+).
The results are reported in Table~\ref{tab:single_quantitative_results}.
In addition to the translational and rotational accuracy, we also report the localization success rate (s.r.).
We define a success as a final lateral error below 0.5m and a yaw angle error below 2.5$^{\circ}$, which is sufficient to initialize a localization system. Only successful cases contribute to accuracy evaluation. We do not report longitudinal pose errors, since this direction is not constrained by the lanes in single images. 

\begin{table}[t]
     \caption{Single image localization success rate (s.r.) and mean translational and rotational errors in percent, meters or degrees.}
     \label{tab:single_quantitative_results}
     \centering
     \begin{tabular}{|cc|l|cccccc|}
         \hline
         $\delta(m)$ & $\delta(^{\circ})$ & & s.r.& lat& z & yaw & pitch & roll \\ \hline
         \multirow{2}{*}{$\pm0.5$} & \multirow{2}{*}{$\pm2.5$} & D & 52 & 0.26 & 0.33 & 1.17 & 1.27 & 1.19  \\
          & & D+ & {78} & {0.24} & {0.23} & {1.05} & {1.00} & {0.97}\\
         \hline
         \multirow{2}{*}{$\pm0.75$} & \multirow{2}{*}{$\pm5.0$} & D & 49 & 0.26 &  0.42 & 1.21 & 1.61 & 1.63  \\
          & & D+ & {68} & {0.24} & {0.33} & {1.04} & {1.38} & {1.25} \\
         \hline
         \multirow{2}{*}{$\pm1.0$} & \multirow{2}{*}{$\pm7.5$} & D & 45 & 0.25 & 0.51 & 1.21 & 2.19 & 2.25  \\
          & & D+ & {57} & {0.24} & {0.41} & {1.00} & {1.80} & {1.41} \\
         \hline
     \end{tabular}
\end{table}

Our uncertainty-based method outperforms the method operating directly on the semantic outputs for every single measure. While the lateral errors show comparable results, we observe that the additional redundancy given by the uncertainty-based method resolves ambiguities w.r.t. the height error and the rotational error. Our method also shows a much higher success rate, proving the robustness towards localization errors. 

\subsection{Pose Graph Localization}
We evaluate our pose graph localization approach on the Lyft 5 dataset, which presents challenging urban scenarios with parking zones, occluded lane borders, and a large number of intersections, where lane borders are barely marked out, see~\figref{fig:qualit_lyft}.
The mean lateral and yaw errors of 0.19\,m and 1.3$^\circ$, presented in Table~\ref{tab:pgl_quantitative_results},
show that our method can keep up with the state of the art, which yields errors in the range of 0.1 - 0.3\,m and 1 - 2$^\circ$, respectively~\cite{caselitz2016monocular,zhang2021efficient,yin2018locnet}.
However, these aforementioned methods either utilize a precise LiDAR sensor or memory-intensive maps.

\begin{figure}
\setlength\tabcolsep{1pt}
\centering
\begin{tabular}{c}
 \includegraphics[width=2.8in]{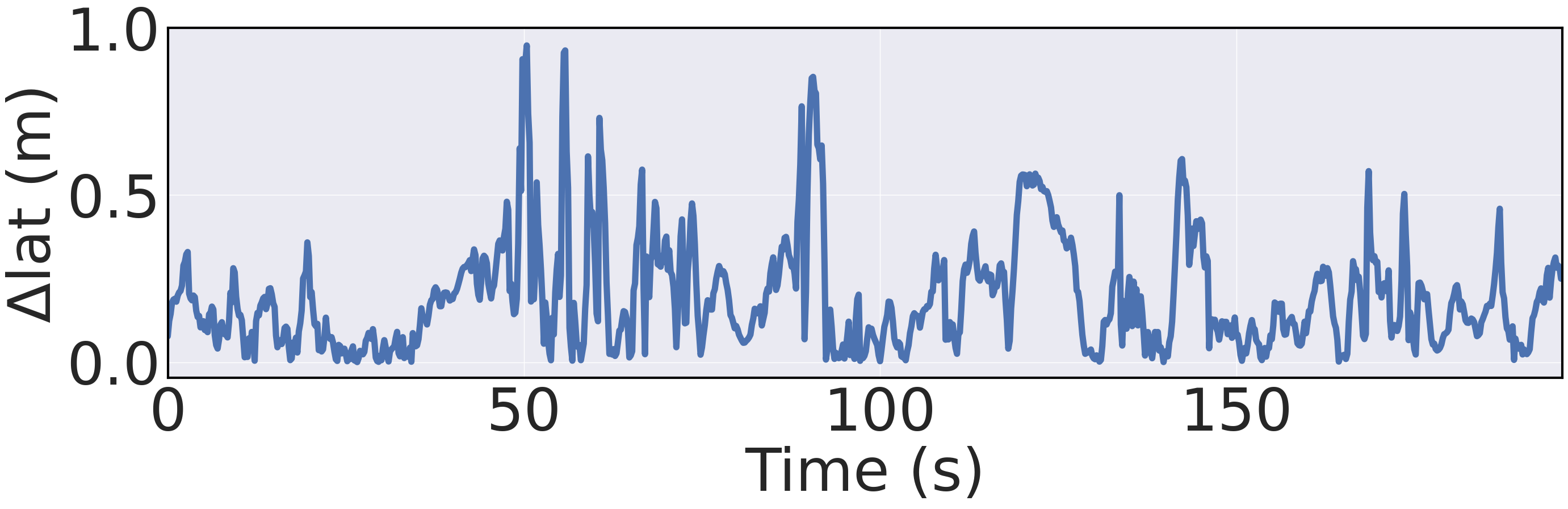} \\
 \includegraphics[width=2.8in]{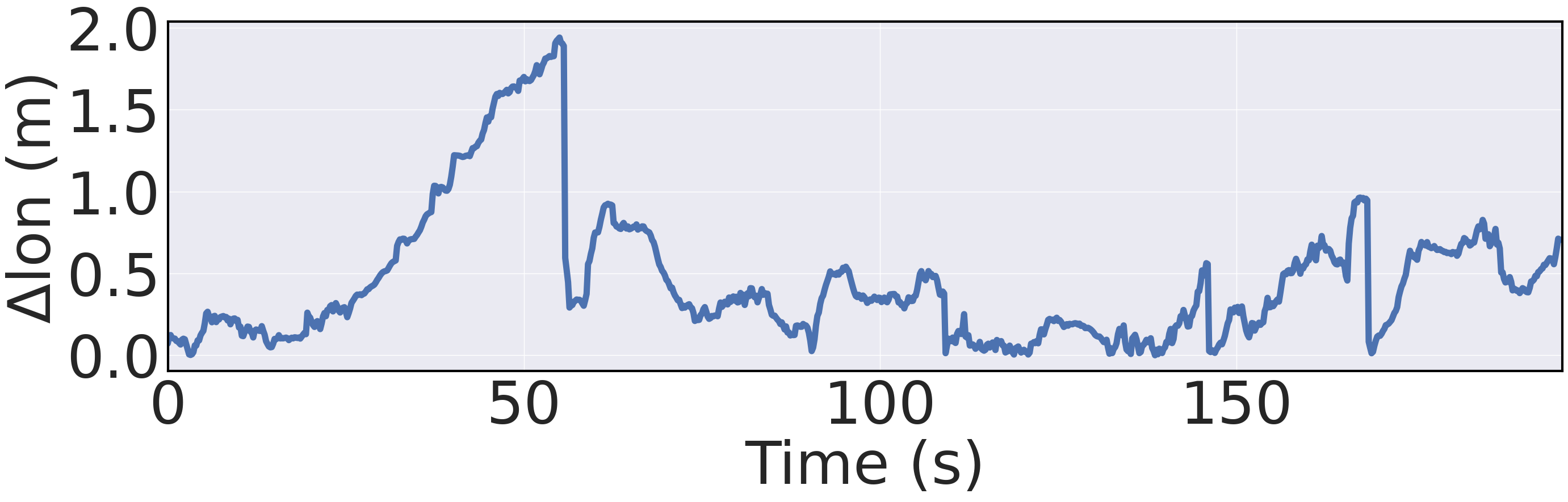} \\
 \includegraphics[width=2.8in]{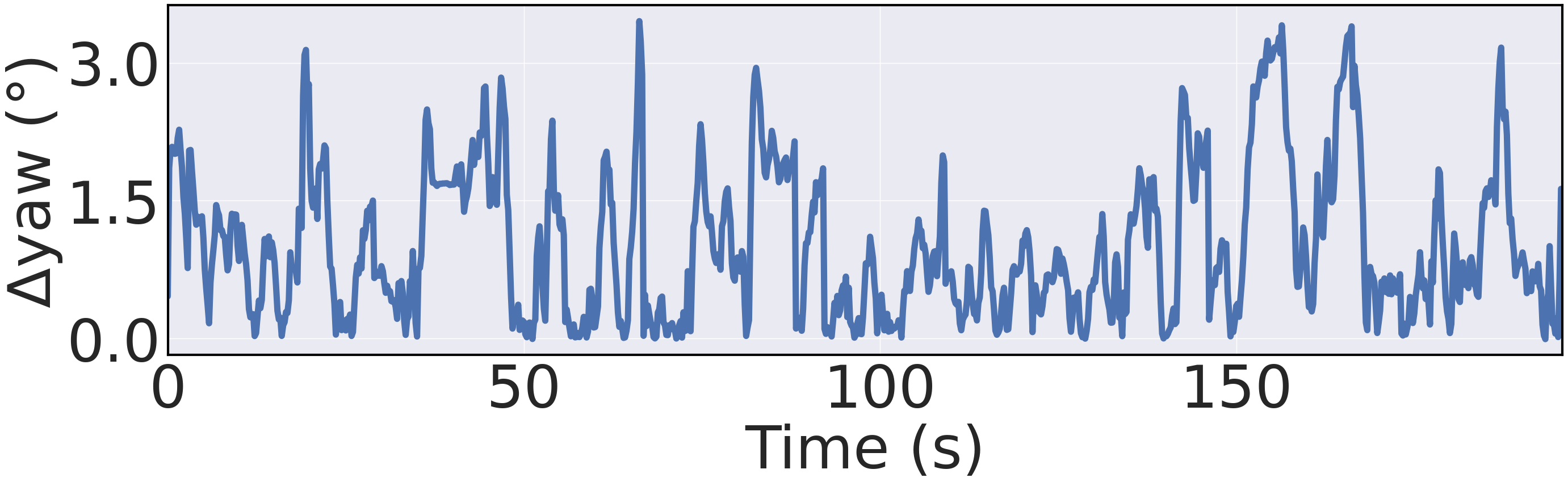} \\
\end{tabular}
\caption{Lateral, longitudinal and yaw errors for our pose graph localization throughout the whole test sequence.}
\label{fig:pgl_quantitative}
\end{figure}

Though our method is laterally accurate on average, complex intersections, and highly populated scenes can degrade the performance due to missing features at intersections and frequent occlusions through parked cars and heavy traffic,
see \figref{fig:pgl_quantitative} (top).
However, our method is able to track the pose even when only a small subset of lane borders is visible.

\begin{table}[t]
     \renewcommand{\arraystretch}{1.3}
     \caption{Pose graph Localization results in meters or degrees relative to the reference provided by Lyft 5.}
     \label{tab:pgl_quantitative_results}
     \centering
     \begin{tabular}{|c|cccccc|}
         \hline
         
         & lon & lat & z & yaw & pitch & roll \\
         \hline
         RMSE & 0.61 & 0.25 & 0.28 & 1.30 & 0.59 &  0.63 \\
         MAE & 0.45&0.19 & 0.17 & 1.01 & 0.45 &  0.42\\\hline
     \end{tabular}
\end{table}

In contrast, the longitudinal pose tends to drift in long driving sequences due to the sparsity of longitudinal cues, see \figref{fig:pgl_quantitative} (middle). It is noticeable that even from large initial errors, the longitudinal pose converges very fast to feasible solutions as soon as traffic lights, or lane features that can break the symmetry in the longitudinal direction, appear. Though relying only on a small set of traffic lights for constraining the longitudinal pose, our localizer manages to keep the mean longitudinal error impressively small by only using constraints in the image plane.

\section{Conclusion}
In this work, we proposed a novel monocular localization system that incorporates predicted uncertainties into a pose graph optimization framework. The uncertainties help to attain robustness in challenging urban scenarios using only sparse map features. As a crucial part of our approach, we presented a novel multi-task uncertainty estimation method that demonstrated the capability to simultaneously learn meaningful uncertainties for semantic segmentation and object detection in a single pass. 
Even though we only use a camera and a sparse map, we demonstrate through our experiments that our approach performs on par with methods that utilize expensive sensor setups or dense maps. 



\bibliographystyle{IEEEtran}
\bibliography{references.bib}

\end{document}